\documentclass{bmvc2k}

\usepackage{times}
\usepackage{epsfig}
\usepackage{graphicx}
\usepackage{amsmath}
\usepackage{amssymb}
\usepackage{multirow}
\usepackage{mathtools}
\usepackage{booktabs}% http://ctan.org/pkg/booktabs
\usepackage{floatrow}
\usepackage{adjustbox}
\title{PanoMixSwap -- Panorama Mixing via Structural Swapping for Indoor Scene Understanding}

\addauthor{Yu-Cheng Hsieh}{sphinx5912@gapp.nthu.edu.tw}{1}
\addauthor{Cheng Sun}{chengsun@gapp.nthu.edu.tw}{1}
\addauthor{Suraj Dengale}{surajdengale@gapp.nthu.edu.tw}{1}
\addauthor{Min Sun}{sunmin@ee.nthu.edu.tw}{1}

\addinstitution{
 Vision Science Lab\\
 National Tsing Hua University\\
 Hsinchu, Taiwan
}

\runninghead{Student, Prof, Collaborator}{PanoMixSwap}

\def\eg{\emph{e.g}\bmvaOneDot}

\def\etal{\emph{et al}\bmvaOneDot}
\newcommand{\modelname}{PanoMixSwap\xspace}

\newcommand{\Sample}{\textbf{S}}
\newcommand{\Image}{I}
\newcommand{\Mask}{M}
\newcommand{\Layout}{L}

\newcommand{\bgstyle}{\mathrm{bs}}
\newcommand{\furnituresetup}{\mathrm{fs}}
\newcommand{\roomstructure}{\mathrm{rs}}
\newcommand{\styledstructure}{\mathrm{ss}}
\newcommand{\augmented}{\mathrm{aug}}

\newcommand{\Sbs}{\Sample_{\bgstyle}}
\newcommand{\Sfs}{\Sample_{\furnituresetup}}
\newcommand{\Srs}{\Sample_{\roomstructure}}
\newcommand{\Saug}{\Sample_{\augmented}}
\newcommand{\Iss}{\Image_{\styledstructure}}
\newcommand{\Ifs}{\Image_{\furnituresetup}}
\newcommand{\Lrs}{\Layout_{\roomstructure}}
\newcommand{\Lbs}{\Layout_{\bgstyle}}
\newcommand{\Lfs}{\Layout_{\furnituresetup}}
\newcommand{\Mbs}{\Mask_{\bgstyle}}
\newcommand{\Mfs}{\Mask_{\furnituresetup}}

\newcommand{\Ialigned}{\Image_{\furnituresetup\rightarrow\roomstructure}}
\newcommand{\Maligned}{\Mask_{\furnituresetup\rightarrow\roomstructure}}

\newcommand{\SFBlock}{\mathrm{\textbf{StyleFusingBlock}}}
\newcommand{\FFBlock}{\mathrm{\textbf{FurnitureFusingBlock}}}

\newcommand{\rind}{r}
\newcommand{\ea}{a}
\newcommand{\eb}{b}
\newcommand{\srca}{\ea_{\mathrm{src}}}
\newcommand{\srcb}{\eb_{\mathrm{src}}}
\newcommand{\dsta}{\ea_{\mathrm{dst}}}
\newcommand{\dstb}{\eb_{\mathrm{dst}}}

\begin{document}

\maketitle

% \begin{abstract}
% The volume and diversity of training data are critical for modern deep learning-based methods.
% Compared to the huge amount of labelled perspective images, ${360}^{\circ}$ panoramic images fall short in both volume and diversity.
% %, which underscores the importance of data augmentation. 
% In this paper, we propose \modelname -- a novel data augmentation technique specifically designed for indoor panoramic images. \modelname explicitly mixes various background styles, foreground furniture, and room layouts from the existing indoor panorama dataset and generates a diverse set of new panoramic images to enrich the dataset. We first decompose each panoramic image into its constituent parts: background style, foreground furniture, and room layout. Then, we generate an augmented image by mixing these parts from different images, such as the foreground furniture from one image, the background style from another image, and the room structure from a third image. This method yields high diversity since there is a cubical increase in image combinations. We also evaluate the effectiveness of \modelname on two indoor scene understanding tasks: semantic segmentation and layout estimation. Our experiments demonstrate that state-of-the-art methods trained with \modelname outperform their original setting on both tasks consistently.
% \end{abstract}

\begin{abstract}
The volume and diversity of training data are critical for modern deep learning-based methods.
Compared to the massive amount of labeled perspective images, ${360}^{\circ}$ panoramic images fall short in both volume and diversity.
In this paper, we propose \modelname, a novel data augmentation technique specifically designed for indoor panoramic images. \modelname explicitly mixes various background styles, foreground furniture, and room layouts from the existing indoor panorama datasets and generates a diverse set of new panoramic images to enrich the datasets. We first decompose each panoramic image into its constituent parts: background style, foreground furniture, and room layout. Then, we generate an augmented image by mixing these three parts from three different images, such as the foreground furniture from one image, the background style from another image, and the room structure from the third image. Our method yields high diversity since there is a cubical increase in image combinations. We also evaluate the effectiveness of \modelname on two indoor scene understanding tasks: semantic segmentation and layout estimation. Our experiments demonstrate that state-of-the-art methods trained with \modelname outperform their original setting on both tasks consistently. The website for this paper can be found at \textit{\url{https://yuchenghsieh.github.io/PanoMixSwap}}.
\end{abstract}
% Data augmentation techniques are often employed to artificially expand the dataset and enhance the diversity of training samples to improve model learning. 
\section{Introduction}\label{intro}
Panoramic images have become increasingly popular in indoor scene understanding tasks because they provide a comprehensive ${360}^{\circ}$ view of a specific room. With the widespread availability of ${360}^{\circ}$ cameras, generating panoramic images has become more convenient. This inspired the development of various indoor panoramic datasets such as Stanford2D3D~\cite{Stanford2D3D}, Matterport3D~\cite{chang2017matterport3d}, PanoContext~\cite{panocontext} and Structured3D~\cite{Structured3D}, as well as the emergence of related tasks such as semantic segmentation, layout estimation, and depth estimation. These tasks leverage the unique characteristics of indoor panoramic images to enable a more holistic and immersive understanding of indoor environments. 

Despite the availability of indoor panoramic datasets, these images are limited in volume and diversity compared to perspective images. For example, even Stanford2D3D~\cite{Stanford2D3D}, one of the largest real-world indoor panoramic datasets, contains only 1,413 panoramic images across 270 scene layouts. This scarcity of data presents difficulties in training models that require both robustness and accuracy. To address this issue, data augmentation techniques are often employed to artificially expand the dataset and enhance the diversity of training samples, thereby mitigating the effects of limited data availability.

% Data augmentation in panoramic images poses unique challenges compared to traditional image data augmentation methods. The inherent structure and layout of panoramic images must be preserved during augmentation to maintain their panoramic format. Some traditional data augmentation techniques, such as random cropping and free-angle rotation, may not be suitable for panoramic images as they can disrupt the intrinsic structure. This underscores the importance of developing novel and specialized data augmentation techniques for panoramic images, which is an active area of ongoing research. 

Data augmentation in panoramic images poses unique challenges compared to traditional image data augmentation methods since the inherent structure and layout of panoramic images must be preserved during augmentation (\eg for indoor panoramic images, ceilings must be on top of walls and floors). Some traditional data augmentation techniques, such as random cropping and free-angle rotation, may not be suitable for panoramic images as they can disrupt the intrinsic structure. This underscores the importance of developing novel and specialized data augmentation techniques for panoramic images.

Current panoramic augmentations are either traditional methods that can preserve the panoramic formats, such as horizontal rotation and flipping, or methods specifically designed for panoramic images like PanoStretch proposed by Sun \etal~\cite{HorizonNet}. However, these methods only work on a single image, which prevents them from combining the variability in different panoramic images as explored by other augmentation methods for perspective images (\eg MixUp~\cite{mixup}). Therefore, present panoramic augmentation methods have limited capability to generate more diverse images.

% To address the issue of limited diversity in indoor panoramic datasets, we utilize multiple panoramic views to augment data and take advantage of variations in different samples. By using two or more panoramic images, semantic masks, or room layouts, we can generate numerous combinations to diversify our training data. Our approach \modelname, as shown in Fig.~\ref{figure:pipeline}, is inspired by the observation that every indoor panoramic image typically consists of three main parts: the room structure ($i.e.$, layout), style of the background (including the ceiling, floor, and each wall), and the foreground furniture. We use these three main parts from three different indoor panoramic views to create a diverse set of augmented samples. Our method leverages a two-stage network to sequentially fuse the background style and foreground furniture into the chosen room layout. The resulting augmented images exhibit a wide range of diverse outputs while preserving the structure of the original panoramic images. We evaluate the effectiveness of our augmentation on two scene understanding tasks: semantic segmentation and layout estimation. By incorporating \modelname during training, we observe significantly improved performance compared to the original settings.

\begin{figure*}
\includegraphics[scale=.25]{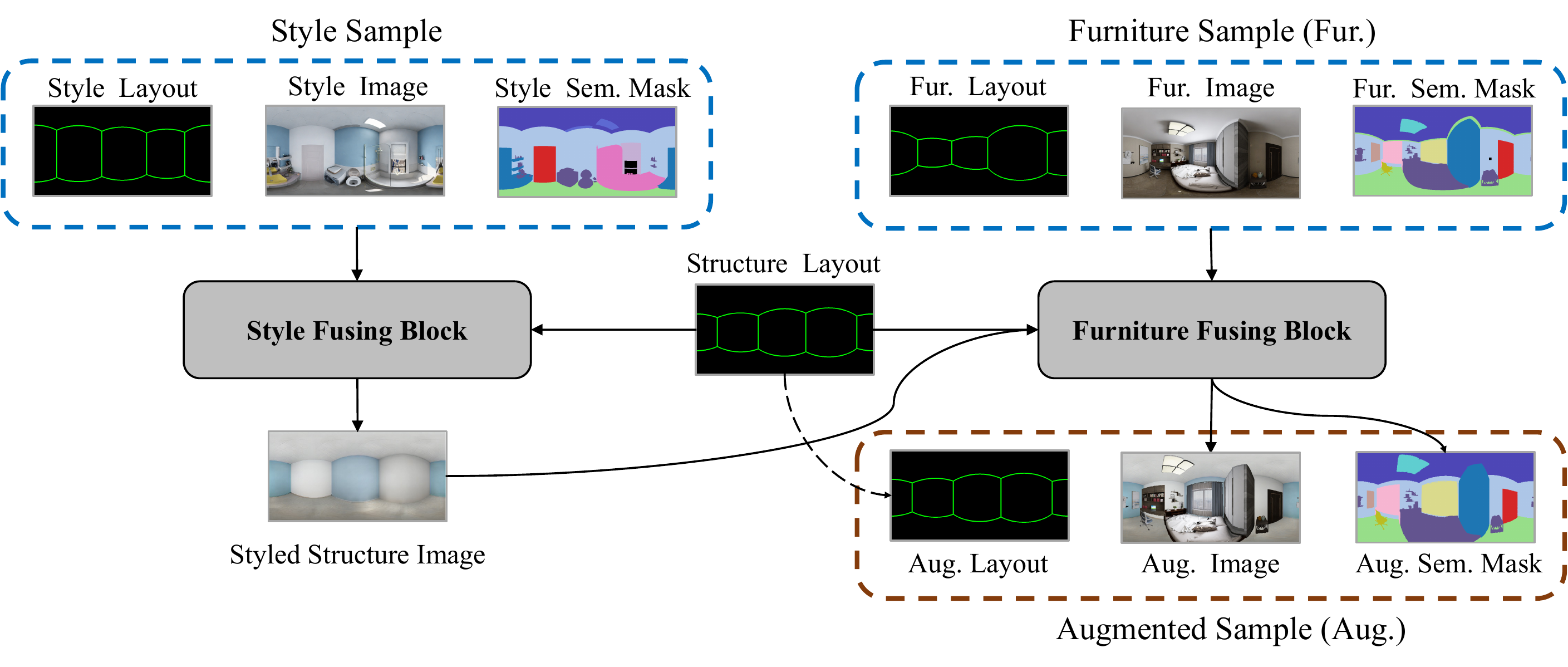}
  \caption{ {\bf Pipeline of \modelname.} \modelname involves three major inputs: style sample, structure layout, and furniture sample. \modelname is composed of two blocks: Style Fusing Block and Furniture Fusing Block. The Style Fusing Block generates a foreground-free styled structure image that fuses the background style from the style image and the room layout from structure layout. Furniture Fusing Block transforms furniture from the furniture image onto the styled structure image to produce the final augmented image and semantic mask.}
  \label{figure:pipeline}
\end{figure*}

To address the limited diversity issue in current panoramic augmentations, we propose a novel panoramic augmentation technique called \modelname, which utilizes multiple panoramic views to augment data and take advantage of variations in different samples. By using two or more panoramic images, semantic masks, and room layouts, we can generate numerous combinations to diversify our training data. \modelname, as shown in Fig.~\ref{figure:pipeline}, is inspired by the observation that every indoor panoramic image typically consists of three main parts: the room structure ($i.e.$, layout), style of the background (including the ceiling, floor, and each wall), and the foreground furniture. We use these three main parts from three different indoor panoramic views to create a diverse set of augmented samples. Our method leverages a two-stage network to sequentially fuse the background style and foreground furniture into the chosen room layout. The resulting augmented images exhibit a wide range of diverse outputs while preserving the structure of the original panoramic images. We evaluate the effectiveness of our augmentation on two scene understanding tasks: semantic segmentation and layout estimation. By incorporating \modelname during training, we observe significantly improved performance compared to the original settings.

Our key contributions to \modelname are summarized below. 
\begin{itemize}
\item We propose a novel data augmentation method \modelname for indoor panoramic images. \modelname generates cubical increased diverse images by mixing three source images while maintaining the structural integrity ($i.e.$, layout). This approach addresses the issue of limited availability in the training data and enhances the variability of the augmented images. 
\item We apply \modelname to two scene understanding tasks, semantic segmentation and layout estimation. \modelname consistently improves results compared to the original training setting.
\end{itemize}

\section{Related Works}
\noindent\textbf{Data Augmentations.}
In the field of computer vision, the size of the dataset plays a crucial role in determining the final performance of the model; hence data augmentation is an important technique for expanding training datasets. Existing data augmentation methods can be categorized into two types: (1) those that use only one training sample to derive one augmented sample and (2) those that use two or more training samples to derive one augmented sample, also called mixup. The first type of augmentation consists of a considerable amount of work, with traditional methods such as random cropping, image mirroring, and color jittering~\cite{AlexNet} commonly used for 2D images, as well as more advanced approaches like AutoAugment~\cite{autoaugment,autoaugment2} and GAN-based methods~\cite{gan1,gan2}. Similarly, for panoramic images, horizontal rotation and flipping techniques and Panostretch~\cite{HorizonNet} introduced in Section~\ref{intro}  are widely used in panoramic-related tasks. On the other hand, the second type of augmentation, $i.e.$, mixup, has been widely studied in 2D image processing, with several works proposing techniques for linearly interpolating two input data points along with their corresponding one-hot labels~\cite{mixup, mixup1, mixup2, mixup3, mixup4, mixup5, mixup6}. For example, Zhang \etal~\cite{mixup6} generate virtual training examples using mixup by linearly interpolating data points and their one-hot labels. Yun \etal~\cite{mixup2} introduce a random-cut rectangular region technique, where a portion of the image is removed and replaced with a patch obtained from another image. Mixup techniques have also been applied in the field of 3D point clouds~\cite{mixup3d1, mixup3d2}. However, to the best of our knowledge, no existing work currently applies the concept of mixup to panoramic images, which serves as a key factor motivating our proposed approach, \modelname.

\noindent\textbf{${\textbf{360}}^{\circ}$ perception.}
The popularity of ${360}^{\circ}$ cameras has recently surged, leading to an increased interest in vision tasks related to panoramic images~\cite{360SD, layoutnet}. Equirectangular projections (ERPs) are commonly used to represent and manipulate the wide field of view captured by these cameras. ERPs allow all captured information to be preserved in a single image. However, they also introduce distortion that can impede the performance of traditional convolution layers designed for perspective images. There has been extensive research on spherical convolution layers~\cite{cohen2018spherical, esteves17_learn_so_equiv_repres_with_spher_cnns, 8953831, NIPS2017_0c74b7f7, DistortionCNN} that are aware of these distortions. To use ${360}^{\circ}$ panoramic images with conventional convolutional neural networks (CNNs) that have a wide range of available pre-trained models, multiple perspective projections are employed to project the image onto multiple planar images. However, this method results in a loss of information due to the projection process, which limits the field of view (FOV). Furthermore, generating planar images from ${360}^{\circ}$ panoramic images requires additional computational resources and time, which increases exponentially with higher-resolution images. To address the problems associated with projection-related works, several newer methods propose different ways of padding~\cite{CubePad, bifuse} and sampling~\cite{tangentImg} image boundaries to remove inconsistencies in panoramic images. The icosahedron mesh~\cite{SpherePHD, Icosahedron} provides a versatile and effective method for representing 3D shapes and scenes in computer vision, particularly for tasks that involve spherical or panoramic data.

\vspace{-2.5mm}
\section{\modelname}\label{approach}
The commonly-used panoramic data augmentations mostly take only one sample as input. %(\eg adding noise, gamma jittering).
However, the diversity of this kind of one-to-one mapping is rather limited.
We propose \modelname to mix three panoramic views into one, which is as clean and high-fidelity as the source views.
Thus, we can generate more diverse training samples which are beyond the conventional panoramic augmentation.

\subsection{Overview}\label{overview}

Let $\Sample$ be a training sample consisting of an RGB image $\Image \in \mathbb{R}^{H \times W \times 3}$, a semantic mask $\Mask \in [0,1]^{H \times W \times C}$ in the form of one-hot vector with $C$ classes, and layout coordinates $\Layout \in \mathbb{R}^{T \times 2 \times 2}$ recording the $T$-walls room corner junctions on floor and ceiling. 
% An output {\it augmented sample} by \modelname is the mix of three samples---a {\it structure sample} $\Srs$, a {\it style sample} $\Sbs$ , and a {\it furniture sample} $\Sfs$.
An output {\it augmented sample} by \modelname is the combination of three main parts from three samples--- room layout structure of {\it structure sample} $\Srs$, background style of {\it style sample} $\Sbs$, and foreground furniture setups of {\it furniture sample} $\Sfs$.
An overview pipeline of \modelname is illustrated in Fig.~\ref{figure:pipeline}.
We first generate a {\it styled structure image} $\Iss$ by mixing the background appearance from $\Sbs$ and the room layout $\Lrs$ from $\Srs$:
\begin{equation}
\label{eq:style_fusing}
    \Iss = \SFBlock(\Sbs, \Lrs) ~,
\end{equation}
where the $\SFBlock$ is detailed in Sec.~\ref{Style Fusing Block}.
We finally can generate the augmented sample $\Saug$ by aligning the furniture setup of $\Sfs$ with the room layout $\Lrs$ and then changing the background style using $\Iss$:
\begin{equation}
\label{eq:furniture_fusing}
    \Saug = \FFBlock(\Sfs, \Lrs, \Iss) ~,
\end{equation}
where $\FFBlock$ is detailed in Sec.~\ref{Furniture Fusing Block}.

\begin{figure*}
\floatbox[{\capbeside\thisfloatsetup{capbesideposition={right,top},capbesidewidth=4.5cm}}]{figure}[\FBwidth]
{\caption{ {\bf Style Fusing Block.} Style Fusing Block is mainly composed of Style Encoder and Style Generator. The Style Encoder is responsible for extracting the embedded style vector for each semantic region of the style image. The Style Generator creates a foreground-free styled structure image by generating the appearance of each semantic region based on its corresponding style embedded 
vector.}\label{figure:SFB}}
{\includegraphics[scale=.25]{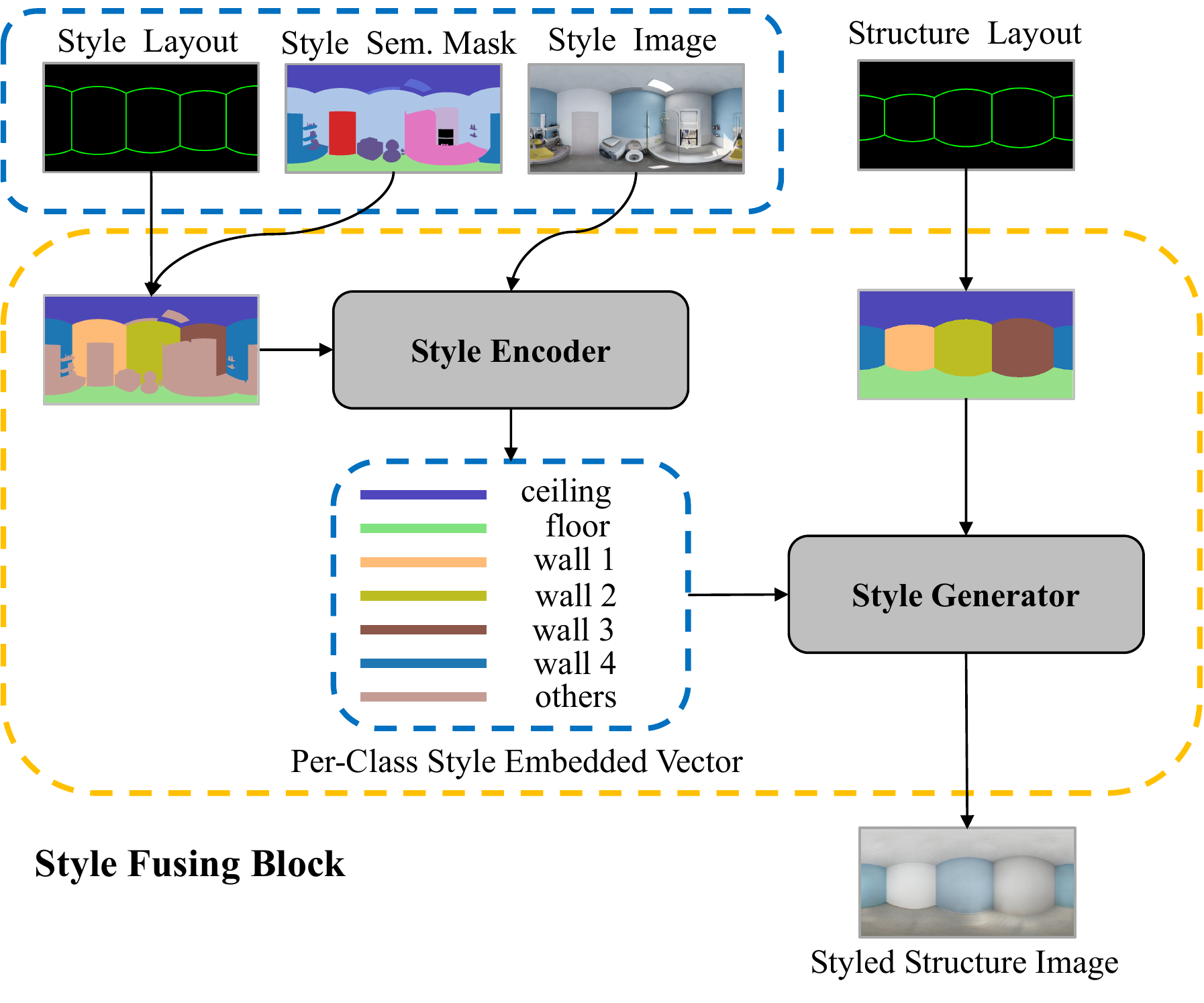}}
\end{figure*}

\subsection{Style Fusing Block}\label{Style Fusing Block}
There are two requirements about the generated styled structure image $\Iss$: \textit{i)} the layout structure should be the same as the room layout $\Lrs$ from structure sample $\Srs$ and \textit{ii)} the background appearance should be similar to the style sample $\Sbs$ with all the furniture removed.
To achieve this, we employ a semantic conditioned generative model, SEAN~\cite{SEAN}.
Specifically, given a content semantic mask, SEAN generates the appearance of each semantic region based on the corresponding semantic region from a reference image.
We use $\Lrs$ to generate the content semantic mask consisting of floor, ceiling, and walls where each wall is assigned a unique class.
The reference semantic mask is generated in the same way using $\Lbs$.
To prevent generating the foreground, the reference semantic mask is further covered by an additional `others' class from the furniture and objects classes in $\Mbs$.
We assume the number of walls is the same in $\Lrs$ and $\Lbs$, so the walls can be one-to-one corresponding.
An overview of the Style Fusing Block is illustrated in (Fig.~\ref{figure:SFB}).
% Please refer to supplementary material for detail about SEAN.

\begin{figure*}
\includegraphics[scale=.176]{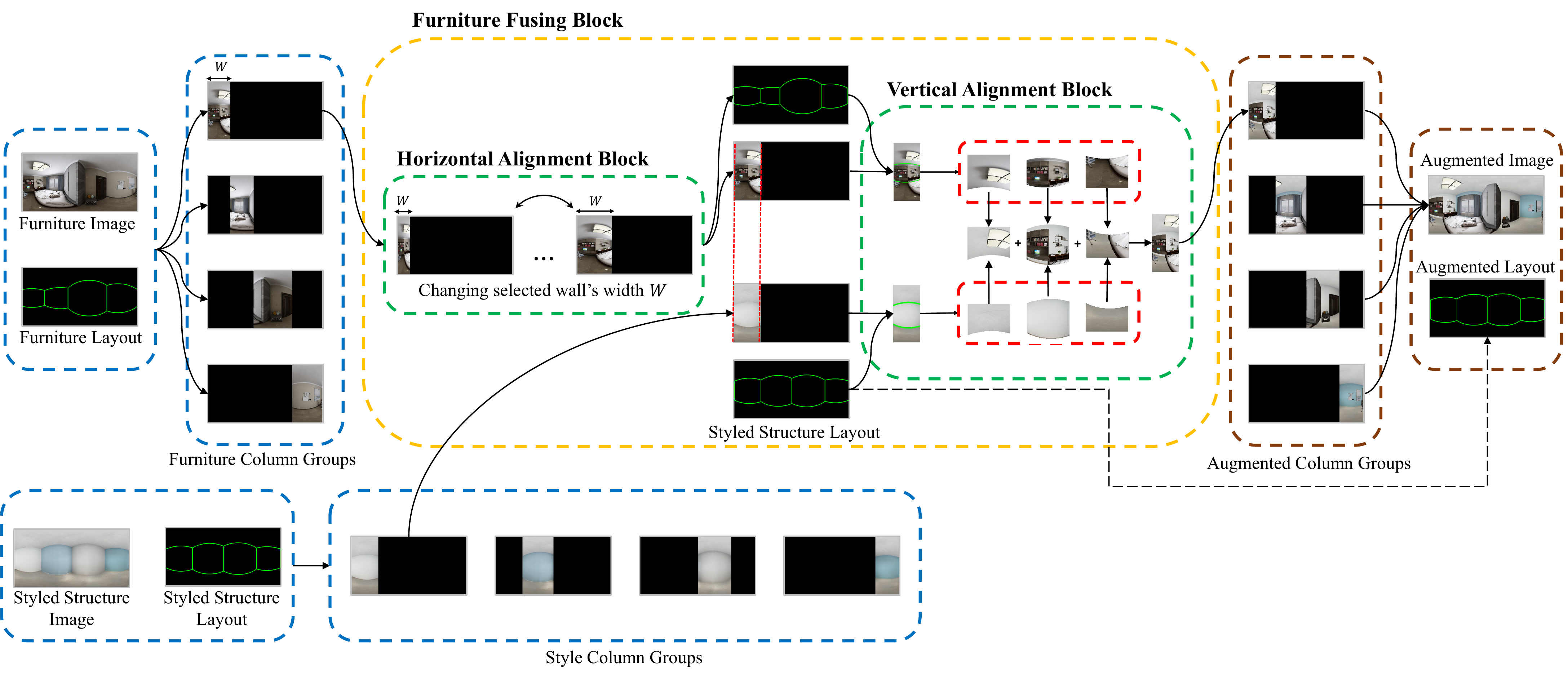}
  \caption{
  {\bf Furniture Fusing Block.} Horizontal Alignment Block takes each furniture column group to produce the width-aligned column group that matches the wall width of the corresponding styled structure column group using PanoStretch~\cite{HorizonNet}. Vertical Alignment Block views both the width-aligned furniture column group and the styled structure column group into ceiling, wall, and floor parts. Then generate the final augmented column group by back warping three parts of the width-aligned furniture column group (denoted aligned furniture column group) to match the same height of three parts from the styled structure column group and replacing background pixel of the styled structure column onto the aligned furniture column group.
  We repeat the process for $T$ times to get the final augmented image.}
  \label{figure:FFB}
\end{figure*}

\subsection{Furniture Fusing Block}\label{Furniture Fusing Block}
The purpose of the Furniture Fusing Block is to fuse the furniture sample $\Sfs$ with the room layout $\Lrs$ and the styled structure image $\Iss$.
To this end, we first align the image $\Ifs$ and the semantic mask $\Mfs$ from their original layout $\Lfs$ to the target layout $\Lrs$.
The aligned image and mask are denoted as $\Ialigned$ and $\Maligned$.
The background pixels of $\Ialigned$ are then replaced by $\Iss$ to change the background style.
% The foreground furniture pixels of $\Ialigned$ then paste onto $\Iss$ to introduce foreground information.
The final {\it augmented sample} is:
\begin{equation} \label{eq:aug_comps}
    \Saug = \{ m\Ialigned + (1-m)\Iss,~ \Maligned,~ \Lrs \} ~,
\end{equation}
where $m$ is the foreground mask computed from $\Maligned$.
Below are the details of the alignment process.

Recap that we assume the number of walls is the same in $\Lfs$ and $\Lrs$, and they are one-to-one corresponding.
We depict the overall process in Fig.~\ref{figure:FFB}.
We first use the wall-wall boundary annotated in $\Lfs$ to split the image columns of $\Ifs$ into multiple {\it image column groups}.
Each image column group is then processed by Horizontal Alignment Block and Vertical Alignment Block sequentially.
In the Horizontal Alignment Block, we use PanoStretch~\cite{HorizonNet} to stretch each image column group from its original width to the corresponding wall width in $\Lrs$.
In the Vertical Alignment Block, we apply backward warping to each image column to align with the ceiling-wall and floor-wall intersection in $\Lrs$.
The source and destination coordinates for the backward warping are computed as follows.
Let $\rind$ be the destination row index of an image column; the source index is computed as
\begin{equation} \label{eq:bw_warp}
    \mathrm{Source}(\rind) = \begin{cases}
        \displaystyle \srca - \alpha(\dsta - r), & \text{if $r < \dsta$} \\
        \displaystyle \srcb + \beta (r - \dstb), & \text{if $r > \dstb$} \\
        \displaystyle \srca + (\srcb - \srca) \frac{(r - \dsta)}{(\dstb - \dsta)}, & \text{otherwise}
    \end{cases} ~,
\end{equation}
where $\ea, \eb$ are the index of the ceiling-wall and floor-wall intersection, $\alpha, \beta$ are hyperparameters. 
The equations in Eq.~\ref{eq:bw_warp} correspond to the warping regions of ceiling, floor, and wall between source and destination. The image column groups are concatenated to form the aligned image $\Ialigned$.
Semantic mask $\Mfs$ is processed in the same way to get $\Maligned$.

\section{Experiments}
We present the implementation details and visualizations of our \modelname in Section~\ref{our Augmentation}.
We showcase the effectiveness of our novel data augmentation technique on indoor ${360}^{\circ}$ semantic segmentation task in Section~\ref{SS} and layout estimation task in Section~\ref{LE}.

\subsection{\modelname}\label{our Augmentation}
\noindent\textbf{Implementation Detail.}
We focus on four-wall indoor panoramic images for simplicity.
To train the encoder-generator model discussed in Sec.~\ref{Style Fusing Block}, we adopt a similar pipeline as proposed in SEAN~\cite{SEAN} for training on both the Structured3D and Stanford2D3D datasets. Specifically, we set the input image size to $H=256$ and $W=512$, use the Adam optimizer with hyperparameters $\beta_1 = 0.5$ and $\beta_2 = 0.999$, and set the learning rate to 2e-4. We use a batch size of 2 and train the model for 60 epochs on a single NVIDIA GTX 1080 Ti GPU. The inference run-time for an image is about 2 seconds, so we apply our augmentation in an offline manner for efficiency.

\begin{figure*}
\includegraphics[scale=.315]{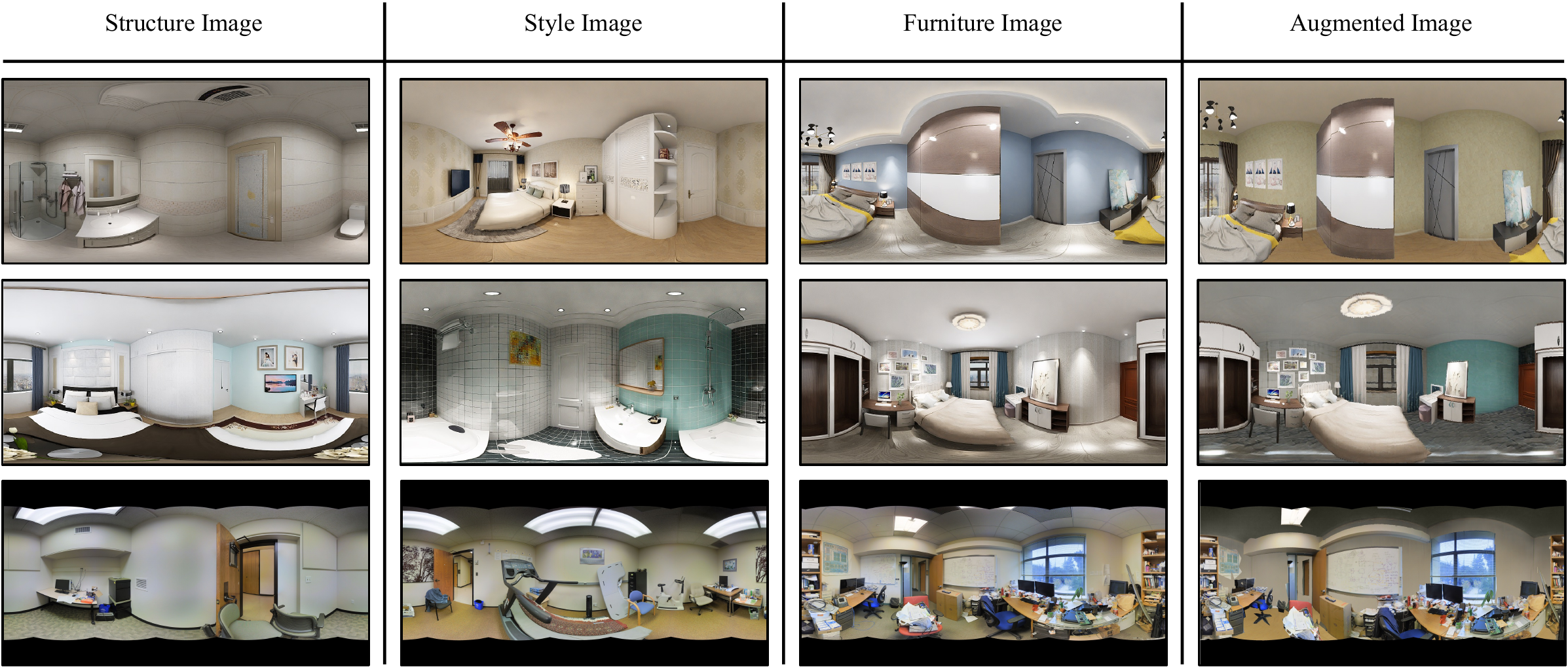}
  \caption{
  {\bf Visualization of the results from our \modelname.}
  % Our novel data augmentation technique takes a structural image, a style image, and a furniture image (first three columns) as input.
  The augmented image (4th column) by our novel \modelname is a fusion of the room layout from the structure image (1st column), the background style from the style image (2nd column), and the furniture from the furniture image (3rd column). 
  The images in the 1st and 2nd rows are from Structured3D~\cite{Structured3D} while the images from the 3rd row are from Stanford2D3D~\cite{Stanford2D3D}.
  }
  \label{figure:qualitative result1}
\end{figure*}

%\subsubsection{Visualizations}
\noindent\textbf{Visualizations.}
 We illustrate the inputs and outputs of \modelname in Fig.~\ref{figure:qualitative result1}.
 Our method can generate a high-quality image by incorporating the background style, room layout structure, and furniture information from three different input samples.
 We use high-quality augmented images to enrich the training set of different tasks.
 For instance, semantic segmentation training data can now be augmented to different room structures and background styles; we can also synthesize different room styles and furniture setups for a given ground-truth room layout.
 
\subsection{Semantic Segmentation}\label{SS}
\noindent\textbf{Model, Dataset and Evaluation.}
In the semantic segmentation task, we use HoHoNet~\cite{HoHoNet} and PanoFormer~\cite{panoformer}, which are two state-of-the-art 360 semantic segmentator. We evaluate \modelname's ability to handle real-world and synthetic data by conducting experiments on two datasets: Stanford2D3D~\cite{Stanford2D3D} and Structured3D~\cite{Structured3D}, whichttps://www.overleaf.com/project/64390116f8940e211f93e500h respectively represent real-world and virtual-world environments.
For Stanford2D3D~\cite{Stanford2D3D}, we use fold 5a and fold 5b for validation and the remaining folds for training following prior works.
As for Structured3D~\cite{Structured3D}, we follow the official training, testing, and validation setting, where there are 3{,}000 scenes for training and 250 scenes for validation, and 250 scenes for testing.
% It's worth noting that we only utilize single RGB images as input in training.
We employ the class-wise mean intersection of union (mIoU) and mean accuracy (mACC) for semantic segmentation evaluation.

\noindent\textbf{Implementation Detail.}
In accordance with the original HoHoNet's setting~\cite{HoHoNet}, we adopt similar implementation settings. For low resolution input, a shallow U-Net with planar CNN is chosen, and the network is trained for 60 epochs on Structured3D~\cite{Structured3D} and 300 epochs on Stanford2D3D~\cite{Stanford2D3D}, using a batch size of 16 and a learning rate of 1e-3 with polynomial decay of factor 0.9. For high resolution input, ResNet-101~\cite{resnet} is used as the backbone, and the network is trained for 60 epochs on both Structured3D~\cite{Structured3D} and Stanford2D3D~\cite{Stanford2D3D}, with a batch size of 4 and a learning rate of 1e-4 with polynomial decay of factor 0.9. For both low resolution and high resolution images, Adam~\cite{Adam} is employed as the optimizer for cross entropy loss.

In the case of PanoFormer~\cite{panoformer}, we use a batch size of 4 and an input resolution of 256*512 to train for 60 epochs. Additionally, Adam~\cite{Adam} is employed as the optimizer for optimizing the cross entropy loss.
To apply \modelname, we first generate an augmented dataset with the same quantity as the original training data and combine the augmented dataset and original training data into a single data set.

\noindent\textbf{Quantitative Results.}
The results of experiments on Stanford2D3D~\cite{Stanford2D3D}, as shown in the upper section of Table~\ref{tabel:Semantic Segmentation}, reveal that the inclusion of our augmentation technique during training leads to significantly higher mIoU and mACC scores on both HoHoNet~\cite{HoHoNet} and PanoFormer~\cite{panoformer} compared to the original work without \modelname, across all models and resolutions. Notably, in high resolution settings, training with \modelname yields a remarkable improvement of 4.02\% in mIoU and 2.43\% in mACC for HoHoNet~\cite{HoHoNet}. 
% We also evaluate the utility of \modelname using PanoFormer~\cite{panoformer} model, and although the improvements in mIoU and mACC are moderate, they are still evident, suggesting that \modelname has a positive impact on the performance of this model as well. 
Based on these compelling results, it is evident that \modelname technique consistently enhances the mIoU and mACC in real-world indoor panoramic scenarios across different models and resolutions.
In addition to real-world scenarios, we also evaluate our augmentation in virtual environment settings using Structured3D dataset~\cite{Structured3D}, as presented in the lower part of Table~\ref{tabel:Semantic Segmentation}. The results demonstrate that training with \modelname leads to higher mIoU and mACC scores in both low and high resolution settings, further substantiating the effectiveness of our technique in virtual indoor panoramic scenarios.

\begin{table}[htbp]
    \centering
    \resizebox{0.99\linewidth}{!}{
    \begin{tabular}{cccccc}
    \toprule
        Dataset & Model & Image Size & \modelname & mIoU(\%) & mACC(\%) \\ \midrule
        \multirow{8}{2cm}{\centering{Stanford2D3D}} &
        \multirow{6}{2cm}{\centering{HoHoNet}}
          & \multirow{2}{2cm}{\centering{$64 \times 128$}} & - & 31.67 & 46.27 \\ 
        ~ & ~ & ~ & \checkmark & \textbf{34.60} & \textbf{47.76} \\ 
        \cline{3-6}
        ~ & ~ & \multirow{2}{2cm}{\centering{$256 \times 512$}} & - & 36.13 & 50.25 \\ 
        ~ & ~ & ~ & \checkmark & \textbf{41.25} & \textbf{52.50} \\ 
        \cline{3-6}
        ~ & ~ & \multirow{2}{2cm}{\centering{$1024 \times 2048$}} & - & 52.00 & 65.00 \\
        ~ & ~ & ~ & \checkmark & \textbf{56.02} & \textbf{67.43} \\
        \cline{2-6}
        ~ & \multirow{2}{2cm}{\centering{PanoFormer}} 
          & \multirow{2}{2cm}{\centering{$256 \times 512$}} & - & 42.20 & 61.03 \\
        ~ & ~ & ~ & \checkmark & \textbf{42.94} & \textbf{62.14} \\ \midrule
        \multirow{6}{2cm}{\centering{Structured3D}} & 
        \multirow{6}{2cm}{\centering{HoHoNet}} 
          & \multirow{2}{2cm}{\centering{$64 \times 128$}} & - & 61.11 & 71.94 \\ 
        ~ & ~ & ~ & \checkmark & \textbf{62.50} & \textbf{73.64} \\ 
        \cline{3-6}
        ~ & ~  & \multirow{2}{2cm}{\centering{$256 \times 512$}} & - & 70.07 & 78.91 \\ 
        ~ & ~ & ~ & \checkmark & \textbf{72.40} & \textbf{81.00} \\ 
        \cline{3-6}
        ~ & ~ & \multirow{2}{2cm}{\centering{$512 \times 1024$}} & - & 80.80 & 87.98 \\ 
        ~ & ~ & ~ & \checkmark & \textbf{81.96} & \textbf{88.52} \\ \bottomrule
    \end{tabular}}
    \caption{
    {\bf Quantitative comparison on semantic segmentation.}
    Our novel \modelname significantly improves two state-of-the-art semantic segmentators, HoHoNet~\cite{HoHoNet} and PanoFormer~\cite{panoformer}, on Stanford2D3D~\cite{Stanford2D3D} and Structured3D~\cite{Structured3D}.
    }
    \label{tabel:Semantic Segmentation}
\end{table}

\subsection{Layout Estimation}\label{LE}
\noindent\textbf{Model, Dataset and Evaluation.}
We utilize HorizonNet~\cite{HorizonNet} and LGT-Net~\cite{lgtnet} to test the effectiveness of \modelname on cuboid layout estimation task, and use the dataset introduced in LayoutNet by Zou \etal~\cite{layoutnet} to estimate cuboid layout. This dataset comprises 514 annotated cuboid room layouts from PanoContext~\cite{panocontext} and 552 annotated cuboid room layouts from Stanford2D3D~\cite{Stanford2D3D}. We follow train/valid/test split in layoutNet~\cite{layoutnet}. For evaluation, we use standard evaluation metrics proposed by Zou \etal~\cite{layoutnet} in cuboid layout estimation, including intersection of union of 3D room layout (3DIoU), corner error (CE), and pixel error (PE). 

\noindent\textbf{Implementation Detail.}
We follow all of the training settings in HorizonNet~\cite{HorizonNet}, which employs a learning rate of 3e-4, and a batch-size of 24 for 300 epochs. In addition, we utilize the training split of Stanford2D3D~\cite{Stanford2D3D} and PanoContext~\cite{panocontext} as training data. 
As for LGT-net~\cite{lgtnet} we train for 1{,}000 epochs with a learning rate of 1e-4 and a batch-size of 6. 
We follow the combined dataset scheme suggested by Zou \etal~\cite{layoutnet_v2}, which involved using the entire PanoContext~\cite{panocontext} and the training split of Stanford2D3D~\cite{Stanford2D3D} as the training data in LGT-net~\cite{lgtnet}. 
For both HorizonNet~\cite{HorizonNet} and LGT-net~\cite{lgtnet}, we employ Adam optimizer~\cite{Adam} with $\beta_{1} = 0.9$, $\beta_{2} = 0.999$ and PanoStretch~\cite{HorizonNet} during training.
% We use different training data settings for HorizonNet~\cite{HorizonNet} and LGT-net~\cite{lgtnet}. For HorizonNet, we utilize the training split of Stanford2D3D~\cite{Stanford2D3D} and PanoContext~\cite{panocontext}, while for LGT-net, we follow the combined dataset scheme suggested by Zou \etal~\cite{layoutnet_v2}, which involved using the entire PanoContext~\cite{panocontext} and the training split of Stanford2D3D~\cite{Stanford2D3D} as the training data. 
In training with \modelname, We apply image augmentation only to the images in the Stanford2D3D~\cite{Stanford2D3D}, and allocate half of the batch size to augmented data and the other half to training data.

\noindent\textbf{Quantitative Results.}
 Table~\ref{tabel:layout} presents a comparison between the performance of using \modelname during training and the original setting on Stanford2D3D~\cite{Stanford2D3D}. 
 % Additionally, we include results obtained by applying post-processing with DuLa-Net~\cite{dulanet} (referred to as "[w/ Post-proc]"). 
 The results show that utilizing \modelname during training outperforms the original setting in 3DIoU, CE on HorizonNet~\cite{HorizonNet} and 3DIoU, PE on LGT-Net~\cite{lgtnet}. 
 Especially on HorizonNet~\cite{HorizonNet}, training with \modelname yields a significant improvement of 3.1\% in 3DIoU.
 This signifies that \modelname has the capability to diversify the training room style and furniture setup, thereby enhancing the overall performance. 

% Stanford2D3D
% \begin{table}[htbp]
%     \centering
%     \begin{tabular}{cccccc}
%     \toprule
%         Model & \modelname & 3DIoU(\%) & CE(\%) & PE(\%) \\ \midrule
%         \multirow{2}{2cm}{\centering{HorizonNet}} & - & 83.51 & 0.62 & \textbf{1.97} \\
%         ~ & \checkmark & \textbf{86.61} & \textbf{0.61} & 1.99 \\\midrule
%         % \multirow{2}{2cm}{\centering{LGT-Net}} & - & 85.76 & - & - \\
%         % ~ & \checkmark & \textbf{86.75} & - & - \\
%         % \cline{2-5}
%         \multirow{2}{2cm}{\centering{LGT-Net}} & - & 86.03 & 0.63 & 2.11 \\
%         ~ & \checkmark & \textbf{86.96} & 0.63 & \textbf{2.04} \\ \bottomrule
%     \end{tabular}
%     ~\\
%     \label{tabel:layout}
%     \caption{
%     {\bf Quantitative comparison on room layout estimation.}
%     Our \modelname can improve HorizonNet~\cite{HorizonNet} and LGT-Net~\cite{lgtnet} on LayoutNet dataset~\cite{layoutnet}.}
% \end{table}

% Stanford2D3D
\begin{table}[htbp]
    \centering
    \begin{tabular}{cccccc}
    \toprule
        Model & \modelname & 3DIoU(\%) & CE(\%) & PE(\%) \\ \midrule
        \multirow{2}{2cm}{\centering{HorizonNet}} & - & 83.51 & 0.62 & \textbf{1.97} \\
        ~ & \checkmark & \textbf{86.61} & \textbf{0.61} & 1.99 \\\midrule
        % \multirow{2}{2cm}{\centering{LGT-Net}} & - & 85.76 & - & - \\
        % ~ & \checkmark & \textbf{86.75} & - & - \\
        % \cline{2-5}
        \multirow{2}{2cm}{\centering{LGT-Net}} & - & 86.03 & 0.63 & 2.11 \\
        ~ & \checkmark & \textbf{86.96} & 0.63 & \textbf{2.04} \\ \bottomrule
    \end{tabular}
    ~\\
    \label{tabel:layout}
    \caption{
    % \vspace{-0.5mm}
    {\bf Quantitative comparison on cuboid room layout estimation.}
    Our \modelname can improve HorizonNet~\cite{HorizonNet} and LGT-Net~\cite{lgtnet} on LayoutNet dataset~\cite{layoutnet}.}
    % \vspace{-1mm}
\end{table}
\subsection{Comparison Between SOTA Augmentation}\label{LE}
This section provides a comprehensive comparison between \modelname and 360 state-of-the-art data augmentation -- PanoStretch proposed by Sun \etal~\cite{HorizonNet} on semantic segmentation task and layout estimation task. The comparison results of semantic segmentation and layout estimation are shown in Table.~\ref{tabel:semantic segmentation cmp} and Table.~\ref{tabel: layout estimation cmp}, respectively. The results of the above two tables show that utilizing \modelname outperforms PanoStretch in above two tasks.

\begin{table}[htbp]
    \centering
    % \resizebox{0.99\linewidth}{!}{
    \begin{tabular}{cccc}
    \toprule
        PanoStretch & \modelname & mIoU(\%) & mACC(\%) \\ \midrule
        - & - & 52.00 & 65.00\\
        \checkmark & - & 53.63 & 65.06\\
        - & \checkmark & \textbf{56.02} & \textbf{67.43}\\
         \checkmark & \checkmark & 55.91 & 67.03\\
    \bottomrule
    \end{tabular}
    %}
    \caption{\textbf{Quantitative comparison between \modelname and PanoStretch on semantic segmentation task}. We use HoHoNet~\cite{HoHoNet} on the Stanford2D3D dataset~\cite{Stanford2D3D} for comparison.}
    \label{tabel:semantic segmentation cmp}
    \vspace{-1em}
\end{table}

\begin{table}[htbp]
    \centering
    \begin{tabular}{cccccc}
    \toprule
        Model & PanoStretch &  \modelname  & 3DIoU(\%) & CE(\%) & PE(\%)\\ \midrule
        \multirow{3}{*}{\centering{HorizonNet}} & \checkmark & - & 83.88 & 0.63 & 2.00 \\
        ~ & - & \checkmark & 85.15 & 0.62 & 1.98\\
        ~ & \checkmark & \checkmark & \textbf{86.59} & \textbf{0.62} & \textbf{1.94}\\
    \bottomrule
        \multirow{3}{*}{\centering{LGT-Net}} & \checkmark & - & 85.98 & 0.65 & 2.11\\
        ~ & - & \checkmark & 86.60 & \textbf{0.62} & 2.06\\
        ~ & \checkmark & \checkmark & \textbf{86.96} & 0.63 & \textbf{2.04}\\
    \bottomrule
    \end{tabular}
    \caption{\bf{Quantitative comparison between \modelname and 360 PanoStretch across the LayoutNet dataset~\cite{layoutnet} on layout estimation task.}}
    \label{tabel: layout estimation cmp}
    \vspace{-1em}
\end{table}

\vspace{-2em}
\subsection{Qualitative Comparison on Downstream Tasks}
\begin{figure*}
\vspace{-1em}
\includegraphics[scale=.22]{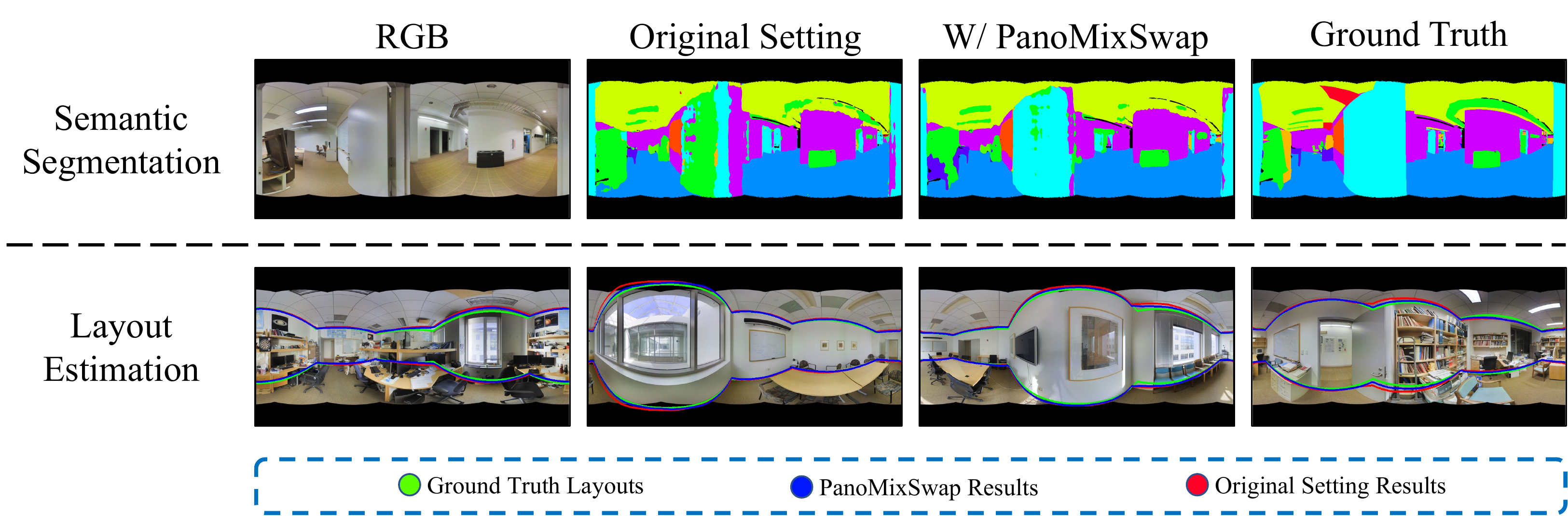}
  \caption{
  {\bf Qualitative comparison on layout estimation and semantic segmentation}
  }
  \label{figure:qualitative simple}
  \vspace{-1em}
\end{figure*}

Fig.~\ref{figure:qualitative simple} presents a qualitative comparison of layout estimation and semantic segmentation. We use HoHoNet~\cite{HoHoNet} as semantic segmentator and HorizonNet~\cite{HorizonNet} as layout estimator. More qualitative results can be found in the supplementary materials.

\vspace{-1em}
\section{Conclusion}
\vspace{-1em}
We present \modelname, a novel data augmentation method for ${360}^{\circ}$ indoor panoramic images. \modelname aims to mix multiple panoramic images to address the issue of data scarcity in panoramic image datasets. Moreover, \modelname introduces an intuitive idea by decomposing a single indoor panoramic image into three distinct parts: foreground furniture, background style, and room layout parts. Then, it mixes multiple panoramic images by swapping these structural parts to generate diverse images. Finally, comprehensive experiments demonstrate that \modelname consistently improves state-of-the-art models on multiple ${360}^{\circ}$ indoor scene understanding tasks.
% To the best of our knowledge, this is the first data augmentation that decomposes a panoramic image into several subdivisions and generates images by combining each subdivision in different image. 

\section*{Acknowledgement}
This work is supported in part by Ministry of Science and Technology of Taiwan (NSTC 111-2634-F-002-022).
We thank National Center for High-performance Computing (NCHC) for computational and storage resource.
We especially thank Chun-Che Wu for providing invaluable guidance for our paper.

\bibliography{egbib}

\begin{thebibliography}{38}
\providecommand{\natexlab}[1]{#1}
\providecommand{\url}[1]{\texttt{#1}}
\expandafter\ifx\csname urlstyle\endcsname\relax
  \providecommand{\doi}[1]{doi: #1}\else
  \providecommand{\doi}{doi: \begingroup \urlstyle{rm}\Url}\fi

\bibitem[360(2020)]{360SD}
360sd-net: 360° stereo depth estimation with learnable cost volume.
\newblock In \emph{2020 IEEE International Conference on Robotics and
  Automation, ICRA 2020}, Proceedings - IEEE International Conference on
  Robotics and Automation, pages 582--588, United States, May 2020. Institute
  of Electrical and Electronics Engineers Inc.
\newblock \doi{10.1109/ICRA40945.2020.9196975}.

\bibitem[Armeni et~al.(2017)Armeni, Sax, Zamir, and Savarese]{Stanford2D3D}
Iro Armeni, Sasha Sax, Amir~R Zamir, and Silvio Savarese.
\newblock Joint 2d-3d-semantic data for indoor scene understanding.
\newblock \emph{arXiv preprint arXiv:1702.01105}, 2017.

\bibitem[Chang et~al.(2017)Chang, Dai, Funkhouser, Halber, Niessner, Savva,
  Song, Zeng, and Zhang]{chang2017matterport3d}
Angel Chang, Angela Dai, Thomas Funkhouser, Maciej Halber, Matthias Niessner,
  Manolis Savva, Shuran Song, Andy Zeng, and Yinda Zhang.
\newblock Matterport3d: Learning from rgb-d data in indoor environments.
\newblock \emph{arXiv preprint arXiv:1709.06158}, 2017.

\bibitem[Chen et~al.(2020)Chen, Hu, Gavves, Mensink, Mettes, Yang, and
  Snoek]{mixup3d1}
Yunlu Chen, Vincent~Tao Hu, Efstratios Gavves, Thomas Mensink, Pascal Mettes,
  Pengwan Yang, and Cees~GM Snoek.
\newblock Pointmixup: Augmentation for point clouds.
\newblock In \emph{Computer Vision--ECCV 2020: 16th European Conference,
  Glasgow, UK, August 23--28, 2020, Proceedings, Part III 16}, pages 330--345.
  Springer, 2020.

\bibitem[Cheng et~al.(2018)Cheng, Chao, Dong, Wen, Liu, and Sun]{CubePad}
Hsien-Tzu Cheng, Chun-Hung Chao, Jin-Dong Dong, Hao-Kai Wen, Tyng-Luh Liu, and
  Min Sun.
\newblock Cube padding for weakly-supervised saliency prediction in 360°
  videos.
\newblock In \emph{2018 IEEE/CVF Conference on Computer Vision and Pattern
  Recognition}, pages 1420--1429, 2018.
\newblock \doi{10.1109/CVPR.2018.00154}.

\bibitem[Cohen et~al.(2018)Cohen, Geiger, Koehler, and
  Welling]{cohen2018spherical}
Taco~S. Cohen, Mario Geiger, Jonas Koehler, and Max Welling.
\newblock Spherical cnns, 2018.

\bibitem[Cubuk et~al.(2019)Cubuk, Zoph, Mane, Vasudevan, and Le]{autoaugment}
Ekin~D Cubuk, Barret Zoph, Dandelion Mane, Vijay Vasudevan, and Quoc~V Le.
\newblock Autoaugment: Learning augmentation strategies from data.
\newblock In \emph{Proceedings of the IEEE/CVF conference on computer vision
  and pattern recognition}, pages 113--123, 2019.

\bibitem[Cubuk et~al.(2020)Cubuk, Zoph, Shlens, and Le]{autoaugment2}
Ekin~D Cubuk, Barret Zoph, Jonathon Shlens, and Quoc~V Le.
\newblock Randaugment: Practical automated data augmentation with a reduced
  search space.
\newblock In \emph{Proceedings of the IEEE/CVF conference on computer vision
  and pattern recognition workshops}, pages 702--703, 2020.

\bibitem[Eder et~al.(2020)Eder, Shvets, Lim, and Frahm]{tangentImg}
Marc Eder, Mykhailo Shvets, John Lim, and Jan-Michael Frahm.
\newblock Tangent images for mitigating spherical distortion.
\newblock In \emph{The IEEE/CVF Conference on Computer Vision and Pattern
  Recognition (CVPR)}, June 2020.

\bibitem[Esteves et~al.(2017)Esteves, Allen-Blanchette, Makadia, and
  Daniilidis]{esteves17_learn_so_equiv_repres_with_spher_cnns}
Carlos Esteves, Christine Allen-Blanchette, Ameesh Makadia, and Kostas
  Daniilidis.
\newblock Learning so(3) equivariant representations with spherical cnns.
\newblock \emph{CoRR}, 2017.

\bibitem[Guo et~al.(2019)Guo, Mao, and Zhang]{mixup3}
Hongyu Guo, Yongyi Mao, and Richong Zhang.
\newblock Mixup as locally linear out-of-manifold regularization.
\newblock In \emph{Proceedings of the AAAI Conference on Artificial
  Intelligence}, volume~33, pages 3714--3722, 2019.

\bibitem[He et~al.(2016)He, Zhang, Ren, and Sun]{resnet}
Kaiming He, Xiangyu Zhang, Shaoqing Ren, and Jian Sun.
\newblock Deep residual learning for image recognition.
\newblock In \emph{2016 IEEE Conference on Computer Vision and Pattern
  Recognition (CVPR)}, pages 770--778, 2016.
\newblock \doi{10.1109/CVPR.2016.90}.

\bibitem[Jiang et~al.(2022)Jiang, Xiang, Xu, and Zhao]{lgtnet}
Zhigang Jiang, Zhongzheng Xiang, Jinhua Xu, and Ming Zhao.
\newblock Lgt-net: Indoor panoramic room layout estimation with geometry-aware
  transformer network.
\newblock In \emph{Proceedings of the IEEE/CVF Conference on Computer Vision
  and Pattern Recognition}, pages 1654--1663, 2022.

\bibitem[Kim et~al.(2020)Kim, Choo, and Song]{mixup1}
Jang-Hyun Kim, Wonho Choo, and Hyun~Oh Song.
\newblock Puzzle mix: Exploiting saliency and local statistics for optimal
  mixup.
\newblock In \emph{International Conference on Machine Learning}, pages
  5275--5285. PMLR, 2020.

\bibitem[Kingma and Ba(2014)]{Adam}
Diederik~P Kingma and Jimmy Ba.
\newblock Adam: A method for stochastic optimization.
\newblock \emph{arXiv preprint arXiv:1412.6980}, 2014.

\bibitem[Krizhevsky et~al.(2012)Krizhevsky, Sutskever, and Hinton]{AlexNet}
Alex Krizhevsky, Ilya Sutskever, and Geoffrey~E Hinton.
\newblock Imagenet classification with deep convolutional neural networks.
\newblock In F.~Pereira, C.J. Burges, L.~Bottou, and K.Q. Weinberger, editors,
  \emph{Advances in Neural Information Processing Systems}, volume~25. Curran
  Associates, Inc., 2012.

\bibitem[Lee et~al.(2019)Lee, Jeong, Yun, Cho, and Yoon]{SpherePHD}
Yeonkun Lee, Jaeseok Jeong, Jongseob Yun, Wonjune Cho, and Kuk-Jin Yoon.
\newblock Spherephd: Applying cnns on a spherical polyhedron representation of
  360° images.
\newblock In \emph{2019 IEEE/CVF Conference on Computer Vision and Pattern
  Recognition (CVPR)}, pages 9173--9181, 2019.
\newblock \doi{10.1109/CVPR.2019.00940}.

\bibitem[Shen et~al.(2022)Shen, Lin, Liao, Nie, Zheng, and Zhao]{panoformer}
Zhijie Shen, Chunyu Lin, Kang Liao, Lang Nie, Zishuo Zheng, and Yao Zhao.
\newblock Panoformer: Panorama transformer for indoor ${360}^{\circ}$ depth
  estimation.
\newblock In \emph{Computer Vision--ECCV 2022: 17th European Conference, Tel
  Aviv, Israel, October 23--27, 2022, Proceedings, Part I}, pages 195--211.
  Springer, 2022.

\bibitem[Sixt et~al.(2018)Sixt, Wild, and Landgraf]{gan1}
Leon Sixt, Benjamin Wild, and Tim Landgraf.
\newblock Rendergan: Generating realistic labeled data.
\newblock \emph{Frontiers in Robotics and AI}, 5:\penalty0 66, 2018.

\bibitem[Su and Grauman(2017)]{NIPS2017_0c74b7f7}
Yu-Chuan Su and Kristen Grauman.
\newblock Learning spherical convolution for fast features from 360\textdegree
  imagery.
\newblock In I.~Guyon, U.~Von Luxburg, S.~Bengio, H.~Wallach, R.~Fergus,
  S.~Vishwanathan, and R.~Garnett, editors, \emph{Advances in Neural
  Information Processing Systems}, volume~30. Curran Associates, Inc., 2017.

\bibitem[Su and Grauman(2019)]{8953831}
Yu-Chuan Su and Kristen Grauman.
\newblock Kernel transformer networks for compact spherical convolution.
\newblock In \emph{2019 IEEE/CVF Conference on Computer Vision and Pattern
  Recognition (CVPR)}, pages 9434--9443, 2019.
\newblock \doi{10.1109/CVPR.2019.00967}.

\bibitem[Sun et~al.(2019)Sun, Hsiao, Sun, and Chen]{HorizonNet}
Cheng Sun, Chi{-}Wei Hsiao, Min Sun, and Hwann{-}Tzong Chen.
\newblock Horizonnet: Learning room layout with 1d representation and pano
  stretch data augmentation.
\newblock In \emph{{IEEE} Conference on Computer Vision and Pattern
  Recognition, {CVPR} 2019, Long Beach, CA, USA, June 16-20, 2019}, pages
  1047--1056, 2019.

\bibitem[Sun et~al.(2021)Sun, Sun, and Chen]{HoHoNet}
Cheng Sun, Min Sun, and Hwann{-}Tzong Chen.
\newblock Hohonet: 360 indoor holistic understanding with latent horizontal
  features.
\newblock In \emph{CVPR}, 2021.

\bibitem[Tateno et~al.(2018)Tateno, Navab, and Tombari]{DistortionCNN}
Keisuke Tateno, Nassir Navab, and Federico Tombari.
\newblock Distortion-aware convolutional filters for dense prediction in
  panoramic images.
\newblock In \emph{Proceedings of the European Conference on Computer Vision
  (ECCV)}, September 2018.

\bibitem[Umam et~al.(2022)Umam, Yang, Chuang, Chuang, and Lin]{mixup3d2}
Ardian Umam, Cheng-Kun Yang, Yung-Yu Chuang, Jen-Hui Chuang, and Yen-Yu Lin.
\newblock Point mixswap: Attentional point cloud mixing via swapping matched
  structural divisions.
\newblock In \emph{Computer Vision--ECCV 2022: 17th European Conference, Tel
  Aviv, Israel, October 23--27, 2022, Proceedings, Part XXIX}, pages 596--611.
  Springer, 2022.

\bibitem[Verma et~al.(2019)Verma, Lamb, Beckham, Najafi, Mitliagkas, Lopez-Paz,
  and Bengio]{mixup4}
Vikas Verma, Alex Lamb, Christopher Beckham, Amir Najafi, Ioannis Mitliagkas,
  David Lopez-Paz, and Yoshua Bengio.
\newblock Manifold mixup: Better representations by interpolating hidden
  states.
\newblock In \emph{International conference on machine learning}, pages
  6438--6447. PMLR, 2019.

\bibitem[Wang et~al.(2020)Wang, Yeh, Sun, Chiu, and Tsai]{bifuse}
Fu-En Wang, Yu-Hsuan Yeh, Min Sun, Wei-Chen Chiu, and Yi-Hsuan Tsai.
\newblock Bifuse: Monocular 360 depth estimation via bi-projection fusion.
\newblock In \emph{Proceedings of the IEEE/CVF Conference on Computer Vision
  and Pattern Recognition (CVPR)}, June 2020.

\bibitem[Yoo et~al.(2020)Yoo, Ahn, and Sohn]{mixup5}
Jaejun Yoo, Namhyuk Ahn, and Kyung-Ah Sohn.
\newblock Rethinking data augmentation for image super-resolution: A
  comprehensive analysis and a new strategy.
\newblock In \emph{Proceedings of the IEEE/CVF Conference on Computer Vision
  and Pattern Recognition}, pages 8375--8384, 2020.

\bibitem[Yun et~al.(2019)Yun, Han, Oh, Chun, Choe, and Yoo]{mixup2}
Sangdoo Yun, Dongyoon Han, Seong~Joon Oh, Sanghyuk Chun, Junsuk Choe, and
  Youngjoon Yoo.
\newblock Cutmix: Regularization strategy to train strong classifiers with
  localizable features.
\newblock In \emph{Proceedings of the IEEE/CVF international conference on
  computer vision}, pages 6023--6032, 2019.

\bibitem[Zhang et~al.(2019)Zhang, Liwicki, Smith, and Cipolla]{Icosahedron}
Chao Zhang, Stephan Liwicki, William Smith, and Roberto Cipolla.
\newblock Orientation-aware semantic segmentation on icosahedron spheres.
\newblock In \emph{Proceedings of the IEEE/CVF International Conference on
  Computer Vision (ICCV)}, October 2019.

\bibitem[Zhang et~al.(2016)Zhang, Bengio, Hardt, Recht, and Vinyals]{mixup6}
Chiyuan Zhang, Samy Bengio, Moritz Hardt, Benjamin Recht, and Oriol Vinyals.
\newblock Understanding deep learning requires rethinking generalization.
\newblock \emph{arXiv preprint arXiv:1611.03530}, 2016.

\bibitem[Zhang et~al.(2017)Zhang, Cisse, Dauphin, and Lopez-Paz]{mixup}
Hongyi Zhang, Moustapha Cisse, Yann~N Dauphin, and David Lopez-Paz.
\newblock mixup: Beyond empirical risk minimization.
\newblock \emph{arXiv preprint arXiv:1710.09412}, 2017.

\bibitem[Zhang et~al.(2014)Zhang, Song, Tan, and Xiao]{panocontext}
Yinda Zhang, Shuran Song, Ping Tan, and Jianxiong Xiao.
\newblock Panocontext: A whole-room 3d context model for panoramic scene
  understanding.
\newblock In \emph{Computer Vision--ECCV 2014: 13th European Conference,
  Zurich, Switzerland, September 6-12, 2014, Proceedings, Part VI 13}, pages
  668--686. Springer, 2014.

\bibitem[Zheng et~al.(2020)Zheng, Zhang, Li, Tang, Gao, and Zhou]{Structured3D}
Jia Zheng, Junfei Zhang, Jing Li, Rui Tang, Shenghua Gao, and Zihan Zhou.
\newblock Structured3d: A large photo-realistic dataset for structured 3d
  modeling.
\newblock In \emph{Proceedings of The European Conference on Computer Vision
  (ECCV)}, 2020.

\bibitem[Zhu et~al.(2020)Zhu, Abdal, Qin, and Wonka]{SEAN}
Peihao Zhu, Rameen Abdal, Yipeng Qin, and Peter Wonka.
\newblock Sean: Image synthesis with semantic region-adaptive normalization.
\newblock In \emph{IEEE/CVF Conference on Computer Vision and Pattern
  Recognition (CVPR)}, June 2020.

\bibitem[Zhu et~al.(2018)Zhu, Liu, Li, Wan, and Qin]{gan2}
Xinyue Zhu, Yifan Liu, Jiahong Li, Tao Wan, and Zengchang Qin.
\newblock Emotion classification with data augmentation using generative
  adversarial networks.
\newblock In \emph{Advances in Knowledge Discovery and Data Mining: 22nd
  Pacific-Asia Conference, PAKDD 2018, Melbourne, VIC, Australia, June 3-6,
  2018, Proceedings, Part III 22}, pages 349--360. Springer, 2018.

\bibitem[Zou et~al.(2018)Zou, Colburn, Shan, and Hoiem]{layoutnet}
Chuhang Zou, Alex Colburn, Qi~Shan, and Derek Hoiem.
\newblock Layoutnet: Reconstructing the 3d room layout from a single rgb image.
\newblock In \emph{Proceedings of the IEEE conference on computer vision and
  pattern recognition}, pages 2051--2059, 2018.

\bibitem[Zou et~al.(2021)Zou, Su, Peng, Colburn, Shan, Wonka, Chu, and
  Hoiem]{layoutnet_v2}
Chuhang Zou, Jheng-Wei Su, Chi-Han Peng, Alex Colburn, Qi~Shan, Peter Wonka,
  Hung-Kuo Chu, and Derek Hoiem.
\newblock Manhattan room layout reconstruction from a single ${360}^{\circ}$
  image: A comparative study of state-of-the-art methods.
\newblock \emph{International Journal of Computer Vision}, 129:\penalty0
  1410--1431, 2021.

\end{thebibliography}
\end{document}